\documentclass[letterpaper, 10 pt, conference]{ieeeconf}  
\usepackage{graphicx}
\usepackage{float} 
\usepackage{amsmath}
\usepackage{hyperref} 
\newcommand{\dw}{\boldsymbol{w}}
\newcommand{\dC}{\boldsymbol{C}}
\newcommand{\bF}{\boldsymbol{F}}
\newcommand{\bxi}{\boldsymbol{\xi}}
\newcommand{\bA}{\boldsymbol{A}}
\IEEEoverridecommandlockouts                              

\title{\LARGE \bf
Steering Elongate Multi-legged Robots By Modulating Body Undulation Waves 
}

\author{
    Esteban Flores$^{*,1}$\thanks{*These authors contributed equally to this work.}, 
    Baxi Chong$^{*,2}$, 
    Daniel Soto$^{1}$, 
    Daniel I. Goldman$^{2}$\thanks{$^{1}$E. Flores and D. Soto are with Ground Control Robotics LLC. Email: \texttt{eflores36@gatech.edu}, \textit{gcrobotics314@gmail.com}.}\\
    \thanks{$^{2}$B. Chong and D. I. Goldman are with the School of Physics, Georgia Institute of Technology, Atlanta, GA, 30332 USA. Email: \textit{bchong9@gatech.edu},\textit{daniel.goldman@physics.gatech.edu}}\\
    \thanks{Corresponding author: Daniel I. Goldman}
}
\begin{document}

\maketitle
\thispagestyle{empty}
\pagestyle{empty}

\begin{abstract}
Centipedes exhibit great maneuverability in diverse environments due to their many legs and body-driven control. By leveraging similar morphologies and control strategies, their robotic counterparts also demonstrate effective terrestrial locomotion. However, the success of these multi-legged robots is largely limited to forward locomotion; steering is substantially less studied, in part because of the difficulty in coordinating a high degree-of-freedom robot to follow predictable, planar trajectories. To resolve these challenges, we take inspiration from control schemes based on geometric mechanics(GM) in elongate system's locomotion through highly damped environments. We model the elongate, multi-legged system as a ``terrestrial swimmer" in highly frictional environments and implement steering schemes derived from low-order templates of elongate, limbless systems. We identify an effective turning strategy by superimposing two traveling waves of lateral body undulation and further explore variations of the ``turning wave" to enable a spectrum of arc-following steering primitives. We test our hypothesized modulation scheme on a robophysical model and validate steering trajectories against theoretically predicted displacements.  We then apply our control framework to Ground Control Robotics' elongate multi-legged robot, Major Tom, using these motion primitives to construct planar motion and in closed-loop control on different terrains. Our work creates a systematic framework for controlling these highly mobile devices in the plane using a low-order model based on sequences of body shape changes. 
\end{abstract}

\section{Introduction}

Multi-legged robots can traverse rugged landscapes without the need for extensive sensing due to their many legs, making them appealing for various sectors such as search-and-rescue\cite{itoRubbles}, agriculture\cite{pederson2006agriculturalrobots}, and terrestrial exploration. Unlike few-legged systems where effective locomotion relies on the precise control of each leg, a many-legged system can continue to successfully locomote after individual leg failure or missed contacts because of their morphological redundancy~\cite{chong2023multilegged,chong2022general}. 

\begin{figure}[thpb]
    \centering
    \includegraphics[width=\linewidth]{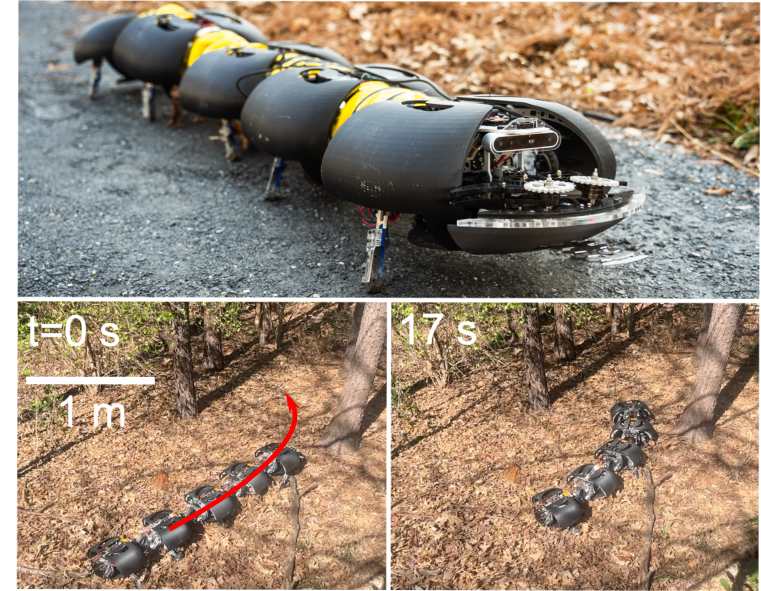}
    \caption{Ground Control Robotics LLC. myriapod robot, Major Tom, autonomously steering around a tree using amplitude modulation of turning wave}
    \label{outside_demo}

\end{figure}

Despite this robustness, whole-body coordination in multi-legged robots (Fig. \ref{outside_demo}) (e.g., body bending, leg stepping, and leg rotation) presents a high degree-of-freedom (DoF) problem that increases in complexity with the number of legs. To reduce the dimensionality of this control problem, schemes based on geometric phase produce effective locomotion by treating elongate locomotors as terrestrial swimmers in highly damped environments\cite{chong2023self}. Using this approach, elongate robot control is simplified to the propagation of a single traveling wave down the body with a coupled leg wave similar to biological systems, achieving robust, agile locomotion using a low-dimensional template that is not limited by number of segments~\cite{rieser2024geometric}\cite{chong2022general}. 
\begin{figure}[thpb]
    \centering
    \includegraphics[width=\linewidth]{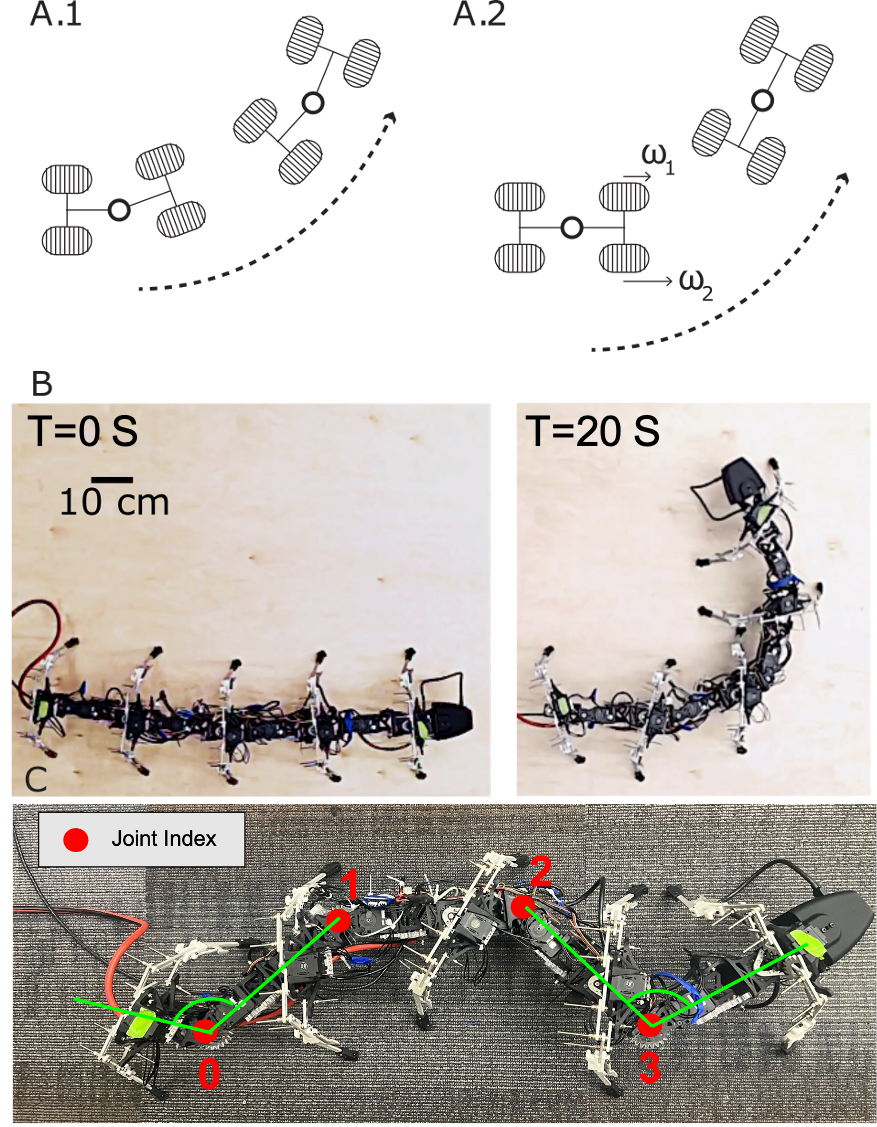} 
    \caption{(A.1) Ackerman steering strategy in a 4 wheeled vehicle utlizing changes to wheel orientation. (A.2) Differential drive steering strategy utilizing rotational velocity imbalance to induce turns on a wheeled vehicle. (B) Centipede robot steering strategy powered by body shape changes similar to Ackerman steering. (C) Horizontal body undulation joint index and angle definition}
    \label{steeringDefinitionFig}

\end{figure}
However, the success of controlling elongate, multi-legged robots using this scheme has been limited to generating forward motion. There exists a significant research gap in producing more complex paths for steering myriapod robots, which severely limits the applicability of these multi-legged systems. Having the ability to follow arcs (i.e., steer) is essential for navigation in cluttered environments and path planning. 

Traditional approaches to steering legged systems take inspiration from the success of wheeled systems that involve an imbalance of angular velocities to create arcs ( Fig.~\ref{steeringDefinitionFig}A.2). The differential drive approach is widely adapted for few-legged robots such as quadrupeds or bipeds. Previous work has shown that careful foot placement planning with rigid bodies can enable quadrupedal turning motions that include considerations for speed, stability, and translation direction \cite{bien1991optimal}\cite{cho1995optimal}. 

However, the differential drive approach is not readily applicable to elongate multi-legged, robots partially because of their relatively simple legs (e.g., mechanical constraints on leg actuation) and their redundant ground reaction force (the differential torque from each leg pair may cancel with each other). Instead, steering strategies generated by body deformation similar to Ackerman steering in wheeled systems ( Fig.~\ref{steeringDefinitionFig}A.1) are more readily applicable to elongate systems with controllable bodies. As a result, steering schemes for myriapod robots leverage mechanically-induced body asymmetry \cite{doi:10.1089/soro.2022.0177} or low-level control of foot contact and body angles, approaches that are difficult to generalize to large numbers of segments\cite{ozkan2021self}. 

To coordinate body deformation that results in predictable motion, we use tools from geometric mechanics (GM) to develop gait sequences and simplify seemingly complex locomotive behavior \cite{batterman2003falling},~\cite{kelly1995geometric,marsden1997geometric,ostrowski1998geometric,shapere1989geometry,wilczek1989geometric,rieser2024geometric}. Building upon success in steering elongate limbless robots~\cite{wang2020omega,wang2022generalized}, we posit similar body-driven turning strategies can be applied to elongate, multi-legged devices because they share a common locomotion framework in environments where damping forces dominate inertial forces~\cite{chong2023self}. 

We propose a simple control strategy that is capable of generating a wide variety of circular arcs by introducing a second traveling wave to the horizontal body undulation (i.e., internal joint angles Fig. \ref{steeringDefinitionFig}.C) while keeping the same coupled leg wave used in forward locomotion \cite{chong2023self}. We implemented this strategy on a robophysical model and in a GM simulation and varied the parameters of the second traveling wave. Notably, the two approaches show close agreement, demonstrating the predictive capability and generality of the two-wave template for undulatory, multi-legged locomotors. In collaboration with Ground Control Robotics LLC, we implemented the steering template on a myriapod robot, Major Tom \cite{groundcontrol2025}, on indoor and outdoor terrain in closed-loop using the onboard force-sensitive antenna (Fig. \ref{outside_demo}). Our results significantly enrich the library of locomotion strategies for elongate multi-legged robots and provide effective tools to navigate around obstacles, paving the way towards all-terrain, highly-capable, elongate multi-legged robots.

\section{Geometric mechanics}
\subsection{Two basis function turning}
\begin{figure}[thpb]
    \centering
    \includegraphics[width=\linewidth]{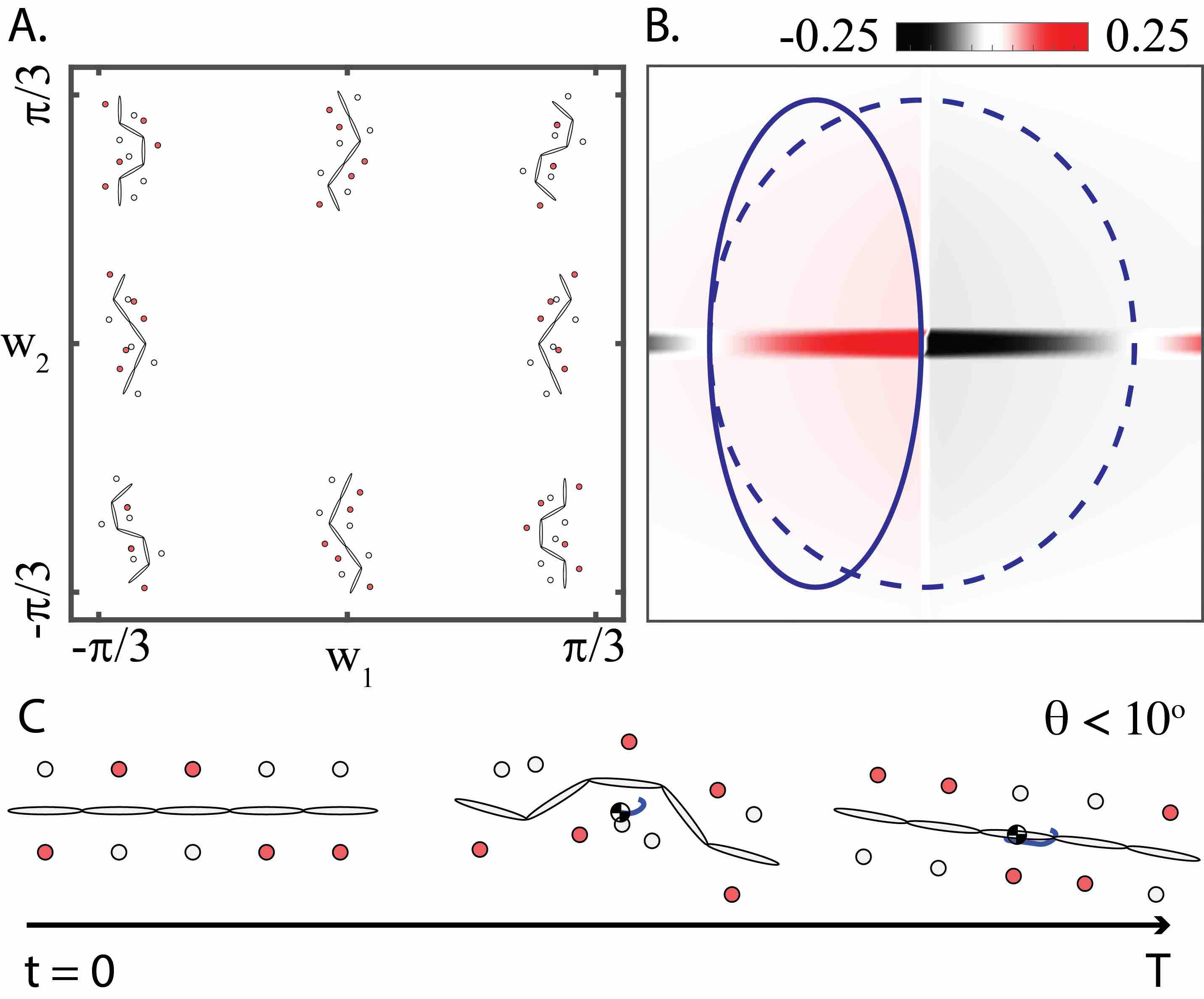}
    \caption{(A) The Euclidean shape space for a multi-legged robot representing a two-wave template. Robot body postures over the shape space is presented. (B) The rotational height functions using the two-wave template. Turning can be modulated by offsetting a gait path from the center (enclosing non-zero surface integral over the height function). Paths in shape space prescribe a series of robot shapes and coupled contact patterns that define a gait. (C) The simulation of the gait leading to net heading angle change less then $10^\circ$
}
    \label{offsetTurn}
\end{figure}
To prescribe horizontal body undulations, we first create a low-dimensional space for multi-legged locomotion. As documented in prior work~\cite{chong2022general,chong2023self}, we assume that the lateral body undulation wave can be prescribed by:

\begin{align}
    \alpha(i,t) &= w_1(t)S_1(i) + w_2(t)S_2(i)  \nonumber \\ 
    S_1 (i) &= \sin{(2\pi k_0 \frac{i}{N})}  \nonumber \\ 
    S_2 (i) &= \cos{(2\pi k_0 \frac{i}{N})} 
\end{align}

\noindent where $\alpha(i,t)$ denotes the yaw angle of $i$-th joint at time $t$ (Fig. \ref{steeringDefinitionFig}C); $S_1(i)$ and $S_2(i)$ are shape basis functions with spatial frequency, or number of waves on the body, $k_0$; $N$ is the number of leg pairs; $w_1(t)$ and $w_2(t)$ are the time series of weights for the corresponding shape basis functions. Here, we define the shape variable as $\dw(t) = [w_{1}(t), w_{2}(t)]$, which then uniquely characterizes the body movement. As discussed in prior work~\cite{chong2022general,chong2023self}, the contact pattern of legs and the leg shoulder angles can also be prescribed by the shape variable $\dw$ (detailed prescription equations can be found in~\cite{chong2022general}). In Fig.\ref{offsetTurn}.A, we illustrated the shape space for a 10-legged robot offset turn which allows for easy comparison of a gait cycle to the two-wave template (Fig \ref{threeWaveTemplate}). 

Perturbation in the shape variable can result in net displacement. The net translation in the plane, SE(2), can be characterized by $\Delta x,\ \Delta y,\ \Delta \theta$  in forward, lateral, and rotational directions, respectively. To better characterize turning/steering in SE(2), we use the following notation to describe the net translation in the plane: $r$, $\Delta \theta$, and $\gamma$, which characterize the steering curvature, rotational translation, and changes in the heading directions, respectively (Fig. \ref{arcDefinition}). Notably, for small perturbations in the shape variables (thus, small displacements), we have: $r\Delta \theta=|[\Delta x, \Delta y]|$, and $\tan(\gamma)=\Delta y / \Delta x$.  

We define the body velocity, $\bxi=[\xi_x, \xi_y, \xi_\theta]$, as the overall locomotor velocity in the forward, lateral, and rotational directions~\cite{murray1994mathematical}. Specifically, we have: $\bxi=\lim_{t\rightarrow0} \frac{[\Delta x,\ \Delta y,\ \Delta \theta]}{dt}$. We can then numerically calculate the body velocity from shape variables ($\dw$) and the shape velocity ($\dot{\dw}$) via net ground reaction forces (GRF) analysis. Here, we model the ground reaction force by Coulomb friction. From geometry, the GRF at each foot can be uniquely expressed as a function of shape variable ($\dw$), shape velocity ($\dot{\dw}$), and body velocity ($\bxi$). Assuming quasi-static motion, we consider the total net force applied to the system is zero at any instant in time:

\begin{equation}\label{eq:forceIntegral}
    \bF=\sum_{i\in I_{\dw}} {\left[\bF^{i}_{\parallel}\left(\bxi,\dw,\dot{\dw}\right)+\bF^{i}_{\perp}\left(\bxi,\dw,\dot{\dw}\right)\right]}=0,
\end{equation}

\noindent where $I_{\dw}$ is the collection of all stance-phase legs, determined by the shape variable $\dw$~\cite{chong2022general}. At a given body shape $\dw$, Eq.(\ref{eq:forceIntegral}) connects the shape velocity $\dot{\dw}$ to the body velocity $\bxi$. Therefore, by the implicit function theorem and the linearization process, we can numerically derive the fundamental equation of motion:

\begin{equation}
    \bxi ~\approx \bA(\dw)\dot{\dw} = \begin{bmatrix}
        \bA^x(\dw) \\ \bA^y(\dw) \\ \bA^\theta (\dw)
    \end{bmatrix}\dot{\dw} ,
\end{equation}

\noindent where $\bA$ is the local connection matrix, $\bA^x, \bA^y, \bA^\theta$ are the three rows of the local connections, respectively. Each row of the local connection matrix can be regarded as a vector field over the shape space, called the connection vector field. In this way, the body velocities in the forward, lateral, and rotational directions are computed as the dot product of connection vector fields and the shape velocity $\dot{\dw}$. 

Consider a gait as a periodic pattern of self-deformation: $\{\partial \phi\ : [w_1(t), w_2(t)], t\in(0,T]\}$, where $T$ is the temporal period. The displacement along the gait path $\partial \phi$ over a cycle can be approximated to the first order by:

\begin{equation}\label{eq:lineintegral}
    \begin{pmatrix} 
        \Delta x \\
        \Delta y \\
        \Delta \theta 
    \end{pmatrix}
    =  \int_{\partial \phi} {\begin{bmatrix}
        \bA^x(\dw) \\ \bA^y(\dw) \\ \bA^\theta (\dw)
    \end{bmatrix}\mathrm{d}\dw}.
\end{equation}

To analyze turning gaits, we have $\Delta \theta = \int_{\partial \phi} \bA^x(\dw) \mathrm{d}\dw$. According to Stokes' Theorem, the line integral along a closed curve $\partial \phi$ is equal to the surface integral of the curl of $\bA^\theta(\dw)$ over the surface enclosed by $\partial \phi$:

\begin{equation}\label{eq:stokes}
    \Delta \theta = \int_{\partial \phi} {\bA^\theta(\dw)\mathrm{d}\dw}=\iint_{\phi} {\nabla\times \bA^\theta(\dw)\mathrm{d}w_1\mathrm{d}w_2},
\end{equation}

\noindent where $\phi$ denotes the surface enclosed by $\partial \phi$. 
The curl of the connection vector field, $\nabla\times \bA^\theta(\dw)$, is referred to as the height function \cite{hatton2015nonconservativity}. With the above derivation, the gait design problem is simplified to drawing a closed path in the shape space. Net displacement over a period can be approximated by the integral of the surface enclosed by the gait path. Hence, we are able to visually identify the optimal gait leading to the largest turning by finding the path with the maximum surface integral. 

We illustrate the height function and an example gait path in Fig.\ref{offsetTurn}B. From the structures of height functions, we notice that (1) the positive and negative volumes are distributed symmetrically about the y-axis ($w_1 = 0$), and (2) turning can be induced if we introduce an offset to the center of the gait path (solid path in Fig.\ref{offsetTurn}B). As documented in~\cite{rieser2024geometric}, adding an offset to the gait path in the shape space is an effective turning strategy for limbless locomotors such as nematode worms and sidewinders. However, we must also note that the magnitude of the rotational height function is small, leading to an less effective turning strategy. To verify our observation, we test the center-shifted path (the solid path in Fig.~\ref{offsetTurn}B) in simulation, and we note that the net turning over a cycle is less than $10^\circ$. We posit that leveraging previous work in limbless locomotion enables more effective steering schemes, specifically through the addition of basis functions.

\subsection{Three basis function turning}
\begin{figure}[thpb]
    \centering
    \includegraphics[width=\linewidth]{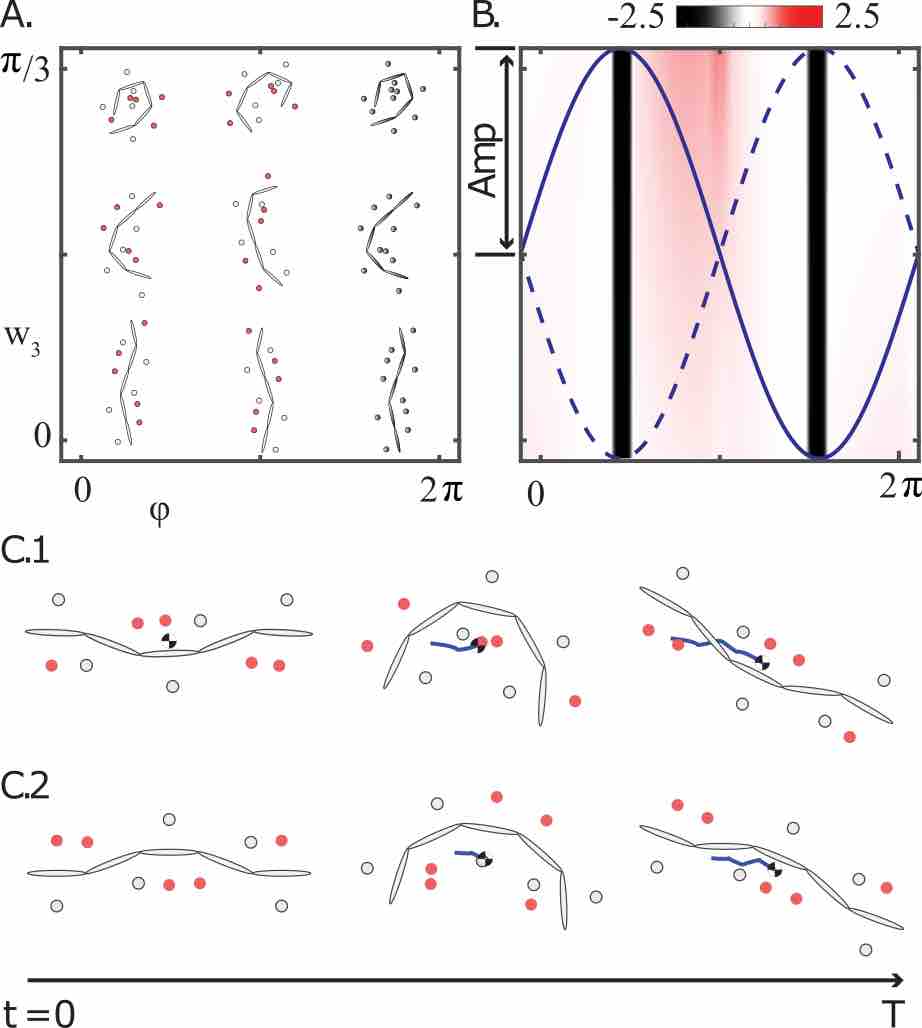} 
    \caption{(A) The cylindrical shape space for a multi-legged robot representing a three-wave template. Y-axis denote the amplitude of the third wave. The x-axis denotes the phase in the first-two waves. (B) The corresponding rotational height functions. We identified two effective steering gaits illustrated in solid and dash curves. The steering gaits can be prescribed by a sinusoidal function with an constant offset. (C) simulation of the two steering gaits. 
}
    \label{threeWaveTemplate}
\end{figure}

We hypothesize that effective turning schemes in nematodes~\cite{wang2020omega,wang2022generalized} may be applicable for multi-legged locomotors because of the similarity in their body driven locomotion through low-inertial regimes. Specifically, we introduce the third basis function:

\begin{align}
    \alpha(i,t) &= w_1(t)S_1(i) + w_2(t)S_2(i) + w_3(t)S_3{i} \nonumber \\ 
    S_3 (i) &= \cos(2\pi k_{1}\frac{i-N/2}{N}) 
\end{align}
\noindent where $k_1$ is the spatial frequency of the third wave. For simplicity, we first consider $k_1=0$ in this section. The third basis function introduces a 3-dimensional shape space, which can present a substantial challenge for directly applying geometric mechanics~\cite{chong2019hierarchical,ramasamy2016soap}. To simplify our analysis, we chose a fixed prescription on the trajectory in $w_1$-$w_2$ space:

\begin{align}
    w_1 &= \pi/6 \cos(\phi) \\
    w_2 &= \pi/6 \sin(\phi),
\end{align}
\noindent where $\phi$ denotes the phase. Thus, we construct a new shape space $\dC = [\phi, \ w_3]$. Notably, $\dC$ has a cylindrical structure \cite{chong2021coordination} because $\phi$ is periodic (i.e., $\phi = 0$ is equivalent to $\phi = 2\pi$). We illustrate the new shape space in Fig. \ref{threeWaveTemplate}. Following similar steps as described in Sec. II.A, we obtain the height function in the new shape space (Fig. \ref{threeWaveTemplate}B). Notably, the new rotational height function is an order of magnitude greater than two basis turning and the gait area (e.g., the solid and dashed line in Fig. \ref{threeWaveTemplate}B) can be calculated as the area underneath the curve, despite the curve not forming a closed surface~\cite{gong2018geometric,lin2020optimizing}. 

We thus identify two effective turning gaits, represented by the solid and dashed curves in Fig.~\ref{threeWaveTemplate}B:

\begin{align}
    w_3 &= A_3 \sin{(\phi)}  \ \ \ or, \\
    w_3 &= A_3 \sin{(\phi+\pi)}
\end{align}

We verify the effectiveness using numerical simulation in Fig.~\ref{threeWaveTemplate}C.  Both gaits can lead to net rotation over $20^\circ$ per cycle, and we choose (Eq.~9) for further analysis. Specifically, we consider a modulation on the amplitude ($A_3$). We hypothesize that the modulation on this amplitude can enable a control over the net rotation  $\Delta \theta$ and turning curvature $r$. To test our hypothesis, we performed the numerical simulation to estimate the net rotation  $\Delta \theta$ and turning curvature $r$ as a function of $A_3$ (numerical simulation details can be found in~\cite{wang2020omega,wang2022generalized,rieser2024geometric}). As a result, we notice an almost linear relationship (Fig.~\ref{results1}.A, red curves) between the amplitude ($A_3$) and the net turning ($\Delta \theta$) or steering curvature ($1/r$).

\section{Experimental Setup}

\begin{figure}[thpb]
    \centering
    \includegraphics[width=\linewidth]{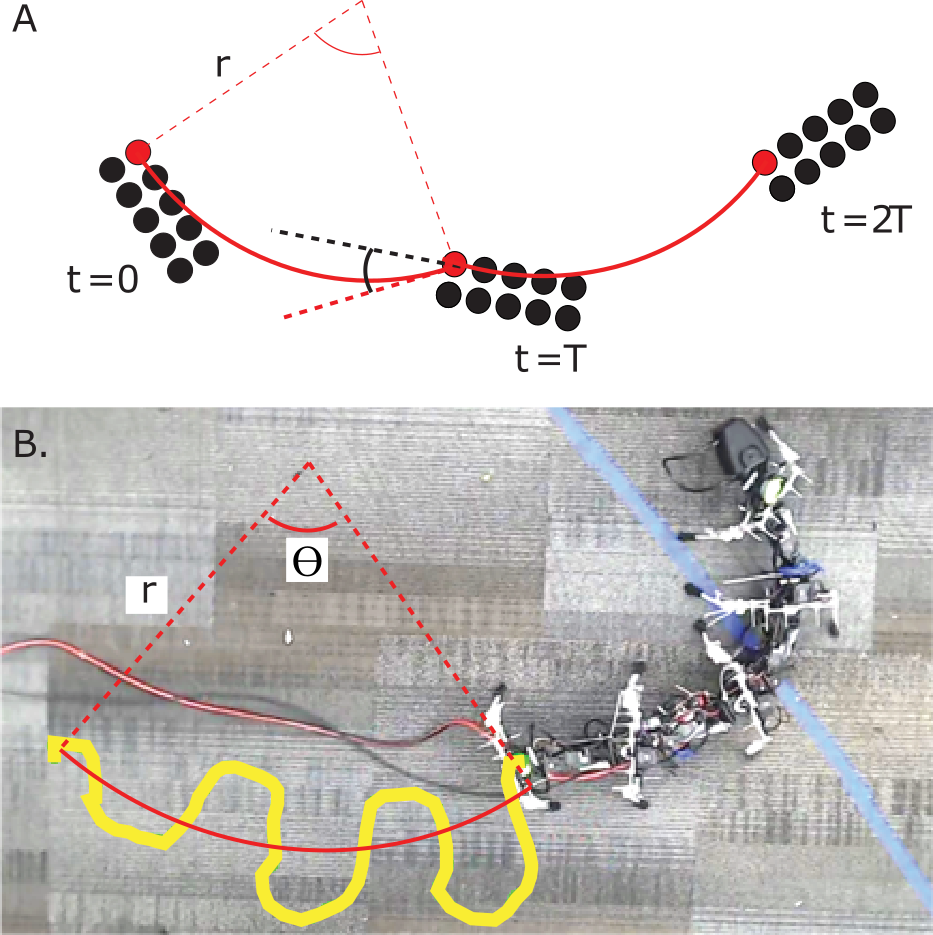}
    \caption{Centipede robot with 5 leg pairs used to test steering control scheme. (A) An illustration of a 10
    -legged robot steering over two cycles. The red curve denotes the trajectory of the last leg on the left. Steering parameters are labeled: steering radius, $R$, heading angle, $\theta$, and orientation change angle from the tangent, $\theta$. (B) Markers are placed on the head and last module of the robot. The yellow curve denotes the marker trajectory over 3 cycles. The red curves are the fitted circle to estimate steering parameters. For clarity, the other marker trajectory was not displayed but was used in fitting the curve. 
}
    \label{arcDefinition}
\end{figure}

To test the possible turning behavior, we built a five-segment multi-legged robophysical model with the ability to control horizontal and vertical body joint angles, leg rotation angles, and leg stepping contacts. Additionally, it has an on-board computer (Raspberry Pi 4), head module with a depth camera (Intel\textsuperscript{\textcopyright} RealSense\textsuperscript{\texttrademark}), and inertial measurement units on each link. Horizontal and vertical body undulation are controlled via Dynamixel 2XL430 servos while leg rotation and stepping motions use Dynamixel XL430 servos. Mechanical gears increase the torque of the body servos while rigid linkages control the stepping motion. There are ball casters on the leg contact to allow for free rotation on the carpet surface and to limit sinkage. 

We define steering paths as circular arcs of fixed radius, \(r\), where the heading angle change, \(\theta\), describes the amount of the arc that is traversed and \(\gamma\) describes the rotation of the robot relative to the tangent of the arc. A planar steering path can be fully defined by these three parameters for each robot cycle (\(t = T\)) as shown in Fig \ref{arcDefinition}A. 

We measured steering trajectories on the robot by tracking markers via an overhead camera (Fig \ref{arcDefinition}B). To reduce noise and variability from the shape of the robot, the average position of two markers was used to capture the bulk motion of the robot, and the vector from tail to head was used to define the pose of the robot body. A circle is fit to the trajectory of the robot after completing 3 cycles to obtain \(r\) while \(\theta\) is obtained by taking the angle between the initial and final pose of the robot. Similarly, \(\gamma\) is calculated as the angle between the tangent of the fitted circle and the pose of the robot. Both \(\theta\) and \(\gamma\) are divided by the number of cycles (in our case, 3) to obtain the average angular displacement per cycle for each trial.

\section{Experimental Results}

\begin{figure}[thpb]
    \centering
    \includegraphics[width=\linewidth]{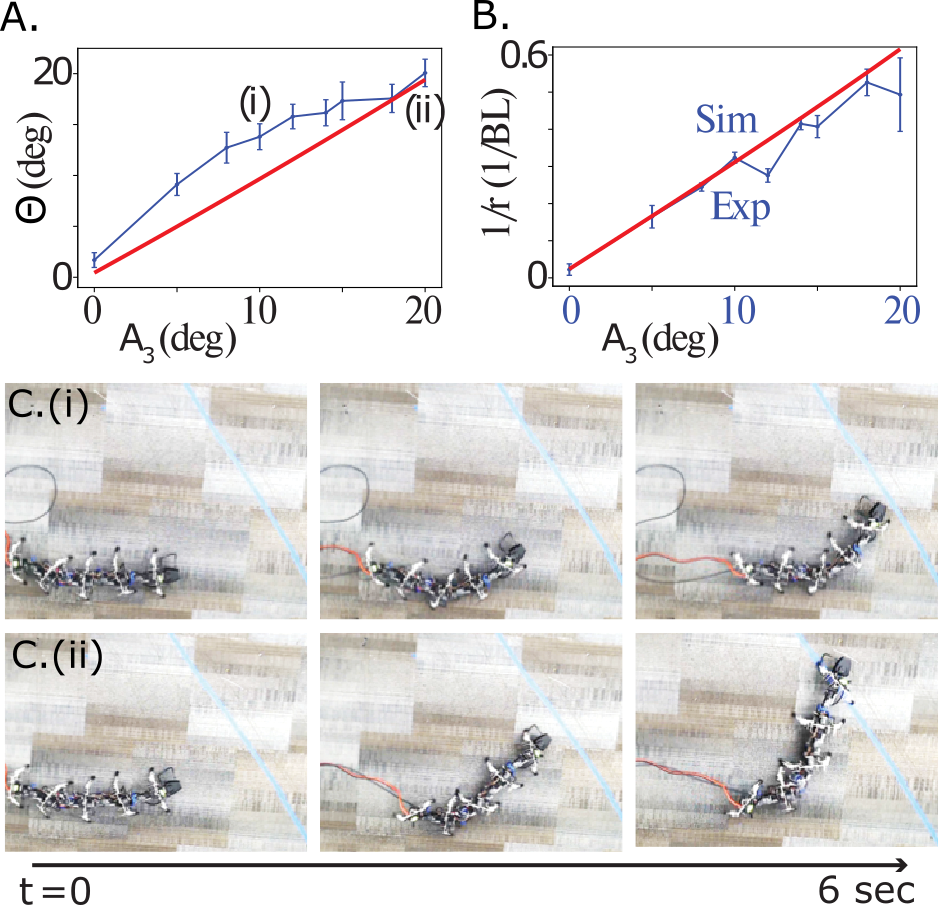} 
    \caption{Experimental and simulation results of (A) heading angle per cycle and (B) curvature per cycle for a range of amplitudes ($A_3$). The points are the average of 5 trials and error bars are 1 standard deviation. (C) By modulating the amplitude of the third wave, different steering paths can be induced on a multi-legged robot. 
}
    \label{results1}
\end{figure}

\subsection{Amplitude Modulation}
We varied the amplitude of \(A_3\) and plot the mean and standard deviation in Figure \ref{results1}. The heading angle change and curvature (\(\frac{1}{r}\)) at amplitudes \(A\in [0 -  20^\circ]\) demonstrate a proportionally increasing relationship with \(A_3\) (Fig \ref{results1}). The resulting average heading angle change shows an increasing trend that appears to be linearly related to \(A_3\) until it begins to saturate after \(10^\circ\) where the slope decreases (Fig. \ref{results1}A). Additionally, due to self collisions at large body joint angles, amplitudes greater than \(20^\circ\) are not possible, yielding a maximum heading angle change of \(20 \pm 1.4\) degrees per cycle. Similarly, the curvature of the steering arcs generally increases with amplitude (Fig.\ref{results1}B), demonstrating that smaller steering radii are achievable with an inversely proportional relationship to amplitude. We observed the greatest orientation change with $A_3 = 20°$ resulting in the robot sweeping 60° over the course of 3 cycles. Interestingly, numerical simulations (Fig.\ref{results1}) for curvature based on GM closely match the linear relationship and bounds for the curvature while the predictions for heading change underestimate a significant portion of the graph. This is potentially due to the discrepancy between the actual ground reaction force between the robot and the carpet and to the assumed Coulomb friction model.

The arc length of the robot path is obtained by multiplying the \(\theta\) by steering radius and it remains relatively constant until \(A_3 = 10^\circ\), where the total path traveled decays as total translation decreases for larger angle rotations with small turning radii. This indicates that, for small amplitude turns, the robot is able to maintain its translational displacement equivalent to purely forward motion (\(0^\circ\)) while its net displacement slowly decays on larger amplitude turns. The rotation angle \(\gamma\) remained constant and close to zero (\(<4^\circ\)) throughout the cycle for each amplitude. While only counterclockwise (CCW) steering data was tracked and reported, clockwise (CW) steering is achieved by negating the shape bases and steering backward requires changing the sign of the first temporal basis function \(w_1\).
\subsection{Spatial Basis Manipulation}
\begin{figure}[thpb]
    \centering
    \includegraphics[width=\linewidth]{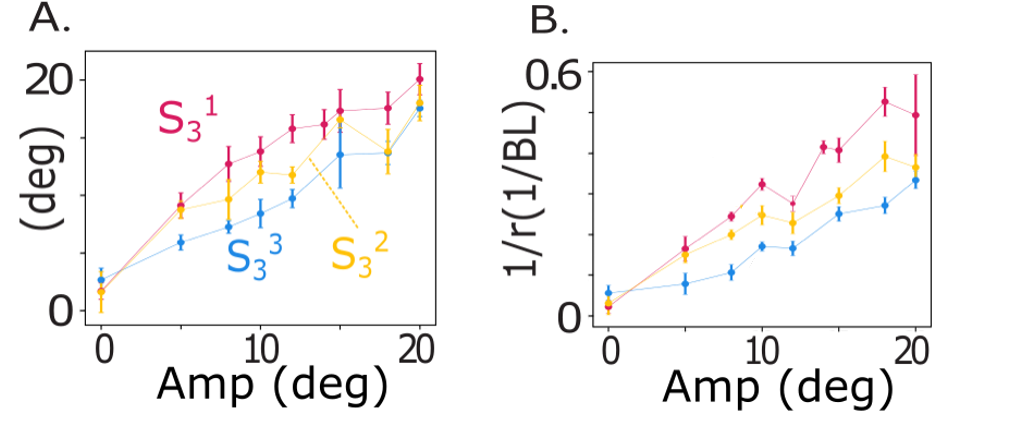} 
    \caption{Experimental results for (A) heading and (B) curvature for three different basis functions \(S_3^2\), \(S_3^2\) and \(S_3^3\)
}
    \label{multipleWaveComparison}
\end{figure}

Our results show that amplitude modulation of a second traveling wave is an effective and simple control for generating steering paths in multi-legged systems. 
However, the third shape basis function, \(S_3\) remained a constant value of [1,1,1,1], denoted as \(S_3^1\), which corresponds to $k_1 = 0$. To investigate the effects of modulating this basis, we conducted trials at two additional static values of \(S_3^2 = [1,.5,.5,1]\) and \(S_3^2 = [1,0,0,1]\), which approximates a spatial frequency of \(k_1 = .2\) and \(k_1 = .4\) respectively (Fig \ref{multipleWaveComparison}). The comparison of the three fixed basis functions demonstrates that manipulation of the third shape basis allows for a wider range of steering trajectories suggesting a second form of control for motion in \(SE(2)\). This data suggests a future avenue for modeling and experiments with a varying \(S_3\) to increase the total steering space in \(SE(2)\) possible for multi-legged robot locomotion.

\section{From Robophysics to Robotics}
\begin{figure}[thpb]
    \centering
    \includegraphics[width=\linewidth]{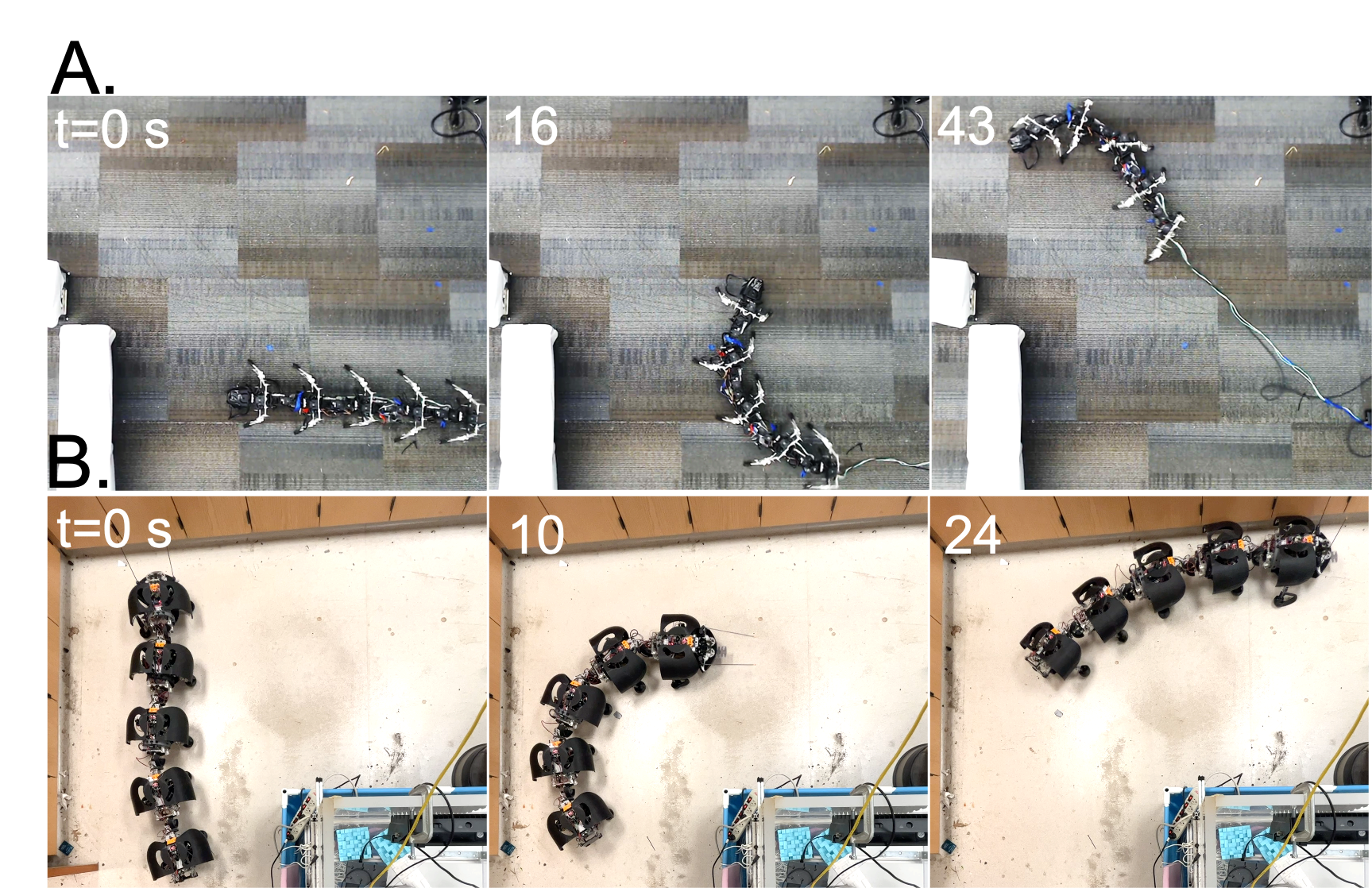}
    \caption{ (A) Robophysical model performing an "S" shaped trajectory around a wall only using amplitude modulation (B) Ground Control Robotics LLC. Major Tom performing wall following algorithm in closed-loop using amplitude modulation steering }
    \label{wall_following}
\end{figure}
Using the empirically derived steering strategies as basic motion primitives, smooth planar motion can be achieved by sequencing these arcs and forward motion together, achieving efficient navigation and obstacle avoidance (Fig. \ref{wall_following}A). The demonstration of the robophysical model turning around a corner in open-loop control, a case where both steering CCW and steering CW are required, shows how different left, right, and forward gait patterns can be sequenced together. Achieving this relatively complex path of sequencing arcs only involves changing the value and sign of \(A_3\) from negative to positive in between cycles while modulating the absolute value dictates the radius of the turn. This demonstrates the power of the dimensionally reduced template for control of high-degree-of-freedom robots and steer in cluttered environments where obstacles must be avoided instead of traversed.

Extending this template to closed-loop control, a test was conducted using Ground Control Robotics LLC.'s (5-link, 1.8 m) Major Tom (Fig. \ref{wall_following}). Using feedback from onboard force-sensing antenna, the robot successfully detected the wall and navigated the corner autonomously using only amplitude modulation of a third basis function. Furthermore, preliminary tests are conducted outside in highly cluttered, rugose environments(Fig. \ref{outside_demo}). Extending the steering templates to other frictional regimes, successful obstacle avoidance maneuvers in closed- and open-loop controls are performed using amplitude modulation, allowing for navigation in scenarios where the robot would otherwise collide with obstacles. 
The extension of the steering template to Ground Control Robotics' Major Tom, approximately five times heavier(10 kg) and 1.5 times as long than the robophysical model,  proves the strength of the dimensionality reduction methods used to control high DOF, undulatory systems while the effectiveness of the steering performance on a variety of terrain (carpet, linoleum, leaves, and dirt) validate the control assumptions of these terrestrial swimmers in low inertia environments. 

\section{Conclusion and Future Work}
Through tools established for limbless locomotors, we effectively extended templates based on geometric mechanics to unlock full planar mobility for undulatory, myriapod-like robots. We present a novel turning scheme that extends previous research in coordinating limbless robots to multi-legged robots by the modulation of an additional traveling wave down the body of the robot which generates a wide variety of steering arcs. Using simulation and robophysical experiments, we found two key variables in controlling steering angles: \(A_3\), the amplitude of the second wave, and \(S_3\), the spatial basis function. While a sufficient range of possible amplitudes were swept, predicting and testing the relationship between spatial frequency modulation and turning performance remains an avenue for future work. Additionally, the motivation behind this work comes from limbless systems, but an investigation in leg dynamics of undulatory, myriapod systems for steering may yield more effective steering schemes and further insights into controlling systems using geometric mechanics.

This scheme builds upon previous work done in shape-based control of multi-legged robots that has proven to be robust across rugged terrain2\cite{chong2023multilegged}. By testing on a variety of terrains (carpet, hard floor, outdoors) and finding close agreement with theoretical predictions, we strengthened the locomotive assumptions in the model for moving in highly damped environments. Furthermore, the predicted turning arcs serve as fundamental motion primitives that are essential for a broad range of locomotion tasks, particularly those involving autonomous navigation in complex environments and obstacle avoidance. Because of the simple amplitude modulation scheme, implementation into closed-loop control involves the manipulation of only one variable for future real-time course correction. We successfully leveraged this fact for basic obstacle avoidance and wall-following algorithms and unlocked potential for work in full trajectory planning for this class of robots. This ability to control planar trajectories serves as a substantial step forward in the mobile applications of multi-legged robot systems, particularly in cluttered and difficult to navigate environments. 



\section*{ACKNOWLEDGMENT}

We would like to thank Erik Teder for his consultation on figures and helpful discussion, and Juntao He for discussion. 
We would like to think the Institute for Robotics and Intelligent Machines for use of the College of Computing basement as a testing space. Additionally, we thank Ground Control Robotics LLC. for use of their robotic platform and support. 
The authors received funding from NSF-Simons Southeast Center for Mathematics and Biology (Simons Foundation SFARI 594594), Army Research Office grant W911NF-11-1-0514, a Dunn Family Professorship, Ground Control Robotics, and a STTR Phase I (2335553) NSF grant.

\clearpage
\bibliographystyle{unsrt}
\bibliography{mainbib}

\begin{thebibliography}{10}

\bibitem{itoRubbles}
Kazuyuki Ito and Shota Kashiwada.
\newblock Proposal of semiautonomous centipede-like robot for rubbles.
\newblock {\em Artificial Life and Robotics}, 19(4):400--405, Dec 2014.

\bibitem{pederson2006agriculturalrobots}
S.~M. Pedersen, S.~Fountas, H.~Have, and B.~S. Blackmore.
\newblock Agricultural robots--system analysis and economic feasibility.
\newblock {\em Precision Agriculture}, 7(4):295--308, 09 2006.
\newblock Copyright - Springer Science+Business Media, LLC 2006; Last updated - 2023-12-03.

\bibitem{chong2023multilegged}
Baxi Chong, Juntao He, Daniel Soto, Tianyu Wang, Daniel Irvine, Grigoriy Blekherman, and Daniel~I Goldman.
\newblock Multilegged matter transport: A framework for locomotion on noisy landscapes.
\newblock {\em Science}, 380(6644):509--515, 2023.

\bibitem{chong2022general}
Baxi Chong, Yasemin~O Aydin, Jennifer~M Rieser, Guillaume Sartoretti, Tianyu Wang, Julian Whitman, Abdul Kaba, Enes Aydin, Ciera McFarland, Kelimar~Diaz Cruz, et~al.
\newblock A general locomotion control framework for multi-legged locomotors.
\newblock {\em Bioinspiration \& Biomimetics}, 17(4):046015, 2022.

\bibitem{chong2023self}
Baxi Chong, Juntao He, Shengkai Li, Eva Erickson, Kelimar Diaz, Tianyu Wang, Daniel Soto, and Daniel~I Goldman.
\newblock Self-propulsion via slipping: Frictional swimming in multilegged locomotors.
\newblock {\em Proceedings of the National Academy of Sciences}, 120(11):e2213698120, 2023.

\bibitem{rieser2024geometric}
Jennifer~M Rieser, Baxi Chong, Chaohui Gong, Henry~C Astley, Perrin~E Schiebel, Kelimar Diaz, Christopher~J Pierce, Hang Lu, Ross~L Hatton, Howie Choset, et~al.
\newblock Geometric phase predicts locomotion performance in undulating living systems across scales.
\newblock {\em Proceedings of the National Academy of Sciences}, 121(24):e2320517121, 2024.

\bibitem{bien1991optimal}
Zeungnam Bien, Myung-Geun Chun, and HS~Son.
\newblock An optimal turning gait for a quadruped walking robot.
\newblock In {\em Proceedings IROS'91: IEEE/RSJ International Workshop on Intelligent Robots and Systems' 91}, pages 1511--1514. IEEE, 1991.

\bibitem{cho1995optimal}
DJ~Cho, JH~Kim, and Dae-Gab Gweon.
\newblock Optimal turning gait of a quadruped walking robot.
\newblock {\em Robotica}, 13(6):559--564, 1995.

\bibitem{doi:10.1089/soro.2022.0177}
Shinya Aoi, Yuki Yabuuchi, Daiki Morozumi, Kota Okamoto, Mau Adachi, Kei Senda, and Kazuo Tsuchiya.
\newblock Maneuverable and efficient locomotion of a myriapod robot with variable body-axis flexibility via instability and bifurcation.
\newblock {\em Soft Robotics}, 10(5):1028--1040, 2023.
\newblock PMID: 37231619.

\bibitem{ozkan2021self}
Yasemin Ozkan-Aydin and Daniel~I Goldman.
\newblock Self-reconfigurable multilegged robot swarms collectively accomplish challenging terradynamic tasks.
\newblock {\em Science Robotics}, 6(56):eabf1628, 2021.

\bibitem{batterman2003falling}
Robert~W Batterman.
\newblock Falling cats, parallel parking, and polarized light.
\newblock {\em Studies in History and Philosophy of Science Part B: Studies in History and Philosophy of Modern Physics}, 34(4):527--557, 2003.

\bibitem{kelly1995geometric}
Scott~D Kelly and Richard~M Murray.
\newblock Geometric phases and robotic locomotion.
\newblock {\em Journal of Robotic Systems}, 12(6):417--431, 1995.

\bibitem{marsden1997geometric}
Jerrold~E Marsden.
\newblock Geometric foundations of motion and control.
\newblock In {\em Motion, Control, and Geometry: Proceedings of a Symposium, Board on Mathematical Science, National Research Council Education, National Academies Press, Washington, DC}, 1997.

\bibitem{ostrowski1998geometric}
Jim Ostrowski and Joel Burdick.
\newblock The geometric mechanics of undulatory robotic locomotion.
\newblock {\em The international journal of robotics research}, 17(7):683--701, 1998.

\bibitem{shapere1989geometry}
Alfred Shapere and Frank Wilczek.
\newblock Geometry of self-propulsion at low reynolds number.
\newblock {\em Journal of Fluid Mechanics}, 198:557--585, 1989.

\bibitem{wilczek1989geometric}
Frank Wilczek and Alfred Shapere.
\newblock {\em \href{http://www.worldscientific.com/worldscibooks/10.1142/0613}{Geometric phases in physics}}, volume~5.
\newblock World Scientific, 1989.

\bibitem{wang2020omega}
Tianyu Wang, Baxi Chong, Kelimar Diaz, Julian Whitman, Hang Lu, Matthew Travers, Daniel~I Goldman, and Howie Choset.
\newblock The omega turn: A biologically-inspired turning strategy for elongated limbless robots.
\newblock In {\em 2020 IEEE/RSJ International Conference on Intelligent Robots and Systems (IROS)}, pages 7766--7771. IEEE, 2020.

\bibitem{wang2022generalized}
Tianyu Wang, Baxi Chong, Yuelin Deng, Ruijie Fu, Howie Choset, and Daniel~I Goldman.
\newblock Generalized omega turn gait enables agile limbless robot turning in complex environments.
\newblock In {\em 2022 International Conference on Robotics and Automation (ICRA)}, pages 01--07. IEEE, 2022.

\bibitem{groundcontrol2025}
{Ground Control Robotics}.
\newblock {Ground Control Robotics Website}.
\newblock \url{https://groundcontrolrobotics.com/}, 2025.
\newblock Accessed: Mar. 3, 2025.

\bibitem{murray1994mathematical}
Richard~M Murray, Zexiang Li, S~Shankar Sastry, and S~Shankara Sastry.
\newblock {\em A mathematical introduction to robotic manipulation}.
\newblock CRC press, 1994.

\bibitem{hatton2015nonconservativity}
Ross~L Hatton and Howie Choset.
\newblock Nonconservativity and noncommutativity in locomotion.
\newblock {\em The European Physical Journal Special Topics}, 224(17-18):3141--3174, 2015.

\bibitem{chong2019hierarchical}
Baxi Chong, Yasemin Ozkan~Aydin, Guillaume Sartoretti, Jennifer~M Rieser, Chaohui Gong, Haosen Xing, Howie Choset, and Daniel~I Goldman.
\newblock A hierarchical geometric framework to design locomotive gaits for highly articulated robots.
\newblock In {\em Robotics: science and systems}, 2019.

\bibitem{ramasamy2016soap}
Suresh Ramasamy and Ross~L Hatton.
\newblock Soap-bubble optimization of gaits.
\newblock In {\em 2016 IEEE 55th Conference on Decision and Control (CDC)}, pages 1056--1062. IEEE, 2016.

\bibitem{chong2021coordination}
Baxi Chong, Yasemin~Ozkan Aydin, Chaohui Gong, Guillaume Sartoretti, Yunjin Wu, Jennifer~M Rieser, Haosen Xing, Perrin~E Schiebel, Jeffery~W Rankin, Krijn~B Michel, et~al.
\newblock Coordination of lateral body bending and leg movements for sprawled posture quadrupedal locomotion.
\newblock {\em The International Journal of Robotics Research}, 40(4-5):747--763, 2021.

\bibitem{gong2018geometric}
Chaohui Gong, Zhongqiang Ren, Julian Whitman, Jaskaran Grover, Baxi Chong, and Howie Choset.
\newblock Geometric motion planning for systems with toroidal and cylindrical shape spaces.
\newblock In {\em Dynamic Systems and Control Conference}, volume 51913, page V003T32A013. American Society of Mechanical Engineers, 2018.

\bibitem{lin2020optimizing}
Bo~Lin, Baxi Chong, Yasemin Ozkan-Aydin, Enes Aydin, Howie Choset, Daniel~I Goldman, and Greg Blekherman.
\newblock Optimizing coordinate choice for locomotion systems with toroidal shape spaces.
\newblock In {\em 2020 IEEE/RSJ International Conference on Intelligent Robots and Systems (IROS)}, pages 7501--7506. IEEE, 2020.

\end{thebibliography}


\end{document}